# Response-state Learning for Transferable Vibrational Spectroscopic Characterization with Electron Prior

Zetong Li[1], Zhuosong Xie[1], Hengyu Fan[1], Jiaao Yu[1], Qiyao Hua[1], Zheng Lu[1], Liming Xu[1], Juanni Wu[1,*], Honglin Li[1,*]

[1]Innovation Center for AI and Drug Discovery, School of Pharmacy, East China Normal University, Shanghai 200062, China.

*To whom correspondence should be addressed.

*Corresponding authors:

Juanni Wu, jnwu@pharm.ecnu.edu.cn

Honglin Li, hlli@hsc.ecnu.edu.cn

Address: 3663 North Zhongshan Road, Shanghai, China. Zip code: 200062

## Abstract

A defining feature of vibrational spectroscopy is that chemically characteristic signals arise predominantly from functional groups and their local chemical and electronic environments, rather than from equivalent contributions across the whole molecule. Consequently, cross-system generalization is governed not simply by molecular size, but by changes in these local environments that reshape vibrational responses. We therefore treat transferable vibrational spectroscopy as a local response-state problem linking structure, electronic environment and spectral signals. We introduce SO(3) Equivariant Neural Kalman Networks (SENK), combining an equivariant backbone for Hessian, dipole-derivative and polarizability-derivative prediction, an Equivariant Neural Kalman bridge for state-dependent calibration, and an NBO-informed electronic prior for bounded, branch-specific calibration. Across molecular benchmarks, drug-like and biomolecular systems, crystal-derived noncovalent assemblies and $\pi$-conjugated oligothiophenes, SENK maintains IR/Raman prediction without system-specific retraining. These results establish local response-state modeling as a physically grounded route to transferable vibrational characterization across molecular and materials-relevant systems.

## Introduction

Molecular vibrational spectroscopy, particularly infrared (IR) absorption and Raman scattering, connects molecular structure with bonding, functional groups, conformations and local chemical environments[1–3]. These fingerprints support molecular identification, structural analysis, reaction monitoring, screening, and environmental and biomolecular applications[1,4–7]. Although recent datasets have expanded reference coverage for small organic molecules and ChEMBL-derived drug-like structures[8,9], they remain limited to the represented chemical domains. Chemically characteristic peak positions and intensities are governed by specific functional groups and by the local chemical and electronic environments that modulate their force constants and response properties. Cross-system generalization is therefore not a simple consequence of molecular size or composition, because changes in hydrogen bonding, conjugation, substitution and polarization can reshape the local vibrational response[3,10,11].

The physical pathway from structure to spectrum proceeds through intermediate response quantities. The mass-weighted Hessian determines vibrational modes and frequencies, whereas dipole and polarizability derivatives determine IR and Raman intensities[1,2]. We refer to these quantities collectively as vibrational response states. Density functional theory, wave-function methods and first-principles dynamics provide accurate spectral assignment and structure-spectrum interpretation[1,2,4,12]. However, repeated electronic-structure calculations, higher-order response evaluations and ab initio trajectories remain costly[4,12–14], limiting high-throughput applications to large molecules, biomolecules, condensed phases and heterogeneous systems[5,7,15]. Harmonic approximations, electronic-structure choices, environmental effects and unusual local environments can further alter peak positions and intensities[3,10,11,16]. Because these local chemical and electronic effects enter the spectrum through the Hessian, dipole derivatives and polarizability derivatives, response states provide a physically direct level at which local environments can be connected to characteristic vibrational signals.

Machine learning has reduced this cost through neural-network potentials and molecular dynamics for IR and Raman simulations[4–7,17–19], as well as graph and attention models for direct structure-to-spectrum prediction[6,10,17,18,20,21]. Equivariant models instead learn the tensorial quantities required by vibrational theory. PaiNN introduced rotationally consistent prediction of dipoles, polarizabilities and spectra[22], and DetaNet extended equivariant tensorial message passing to Hessians and spectroscopic response derivatives[23]. TL-DetaNet transferred full-spectrum prediction to polypeptides and proteins[24]. EnviroDetaNet incorporated multilevel environment information[25], and QMe14S broadened supervision to a 14-element, functional-group-rich chemical space[8]. FIREANN, TNEP, VSpecNN and machine-learning polarizability models have also advanced field-response, dipole, polarizability and Raman prediction in molecules, liquids and solids[13,14,26–32]. These advances establish strong geometry-to-response baselines, while leaving a complementary question: how changes in local chemical environments are expressed in learned response states, how these response variations propagate to characteristic spectral features, and how local electronic information can be incorporated through response-specific calibration.

This study addresses these questions by treating response prediction as a staged state-estimation and calibration problem. Prior work on electronic locality and local vibrational descriptions provides conceptual motivation for using local electronic information to guide transferable response modeling[3,33–35]. Hydrogen bonding, conjugation, inductive and resonance effects, polarization and donor-acceptor interactions modify local force constants and electric-field responses[17,36], and anharmonic couplings may retain a near-sighted organization in larger systems[37]. Natural bond orbital (NBO) theory describes these stereoelectronic interactions through bonds, lone pairs, antibonds, charge redistribution and donor-acceptor coupling[38–40]. These physical ideas motivate an electronic prior that does not replace the learned response representation, but supplies a bounded direction for selected local corrections.

Here we introduce SO(3) Equivariant Neural Kalman Networks (SENK). An equivariant transformer establishes the primary equilibrium-geometry-to-response-tensor mapping. An Equivariant Neural Kalman (ENK) bridge performs state-dependent refinement and supplies an environment-sensitive response-state reliability signal. An NBO-informed electronic-prior (EP) pathway combines consistency-regularized representation learning with guided spectral calibration (GSC) for eligible response branches. We evaluate whether this cascade can maintain an accurate global response baseline, identify localized reliability loss and selectively calibrate response states for which transferable electronic information is available. Evaluation spans QM9S and QMe14S, external drug-like molecules and bioactive peptides, followed by two materials-relevant zero-shot tests: full-spectrum IR/Raman prediction for crystal-derived noncovalent molecular trimers in R-3B69 and conjugation-dependent Raman characterization of π-conjugated oligothiophene semiconductor building blocks against independent dilute-solution measurements[41,42]. Figure 1a summarizes the scientific context and representative application scenarios, whereas Figure 1b presents the SENK architecture.

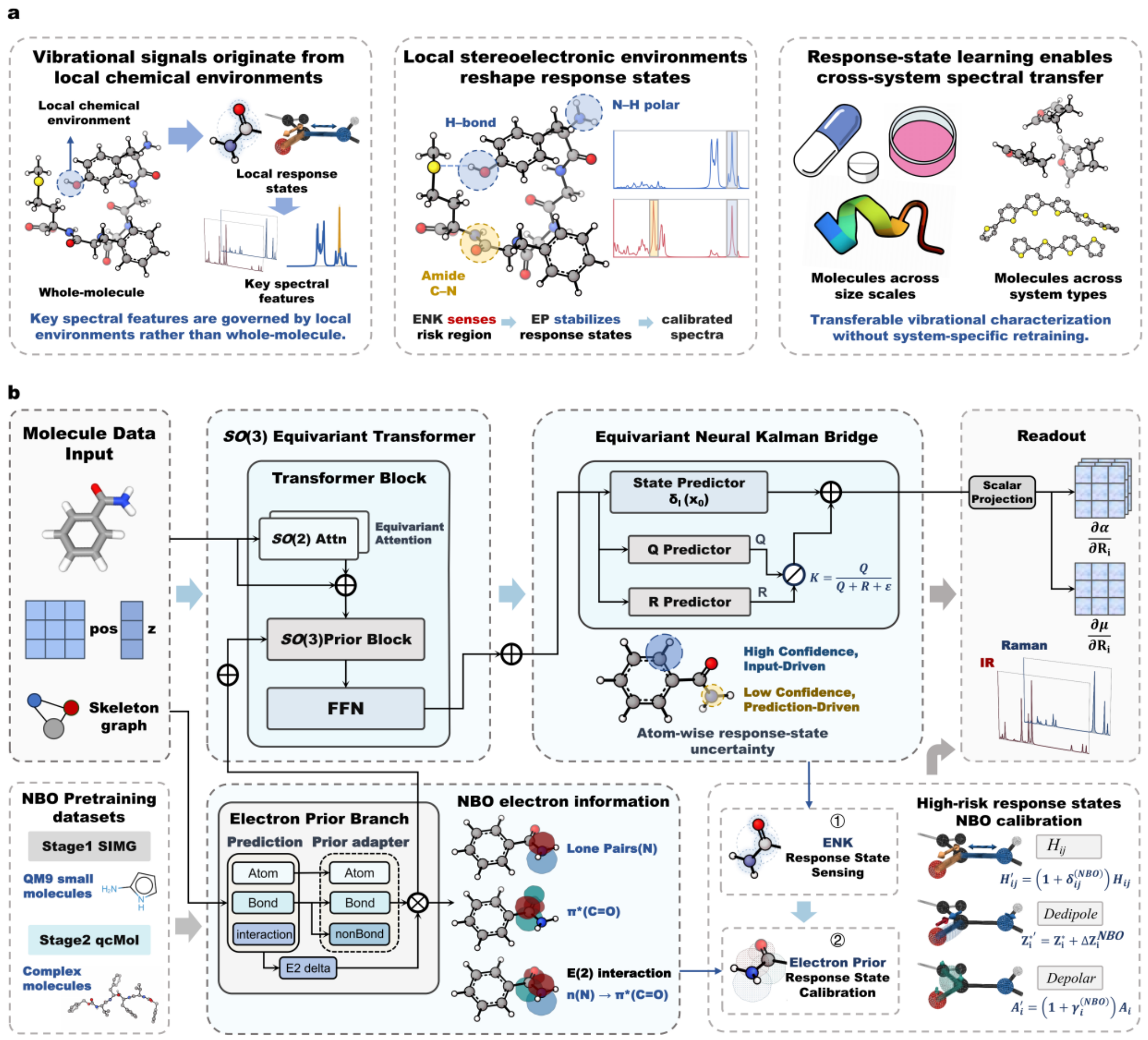

**Figure 1.** Overview of the SENK study and framework. (a) Scientific objective and representative application scenarios of SENK. (b) SENK architecture for response-state uncertainty sensing and selective electron-prior calibration.

## Results

### 1. Performance of SENK in property prediction

Vibrational spectra are determined by a hierarchy of molecular response quantities: the Hessian defines the normal modes and vibrational frequencies, whereas dipole and polarizability derivatives determine the corresponding IR and Raman intensities. We therefore first evaluated whether SENK could establish an accurate geometry-to-response-state mapping before examining its behavior in chemically more complex environments. SENK was benchmarked on QM9S and QMe14S for the combined Hessian matrix, dipole derivatives and polarizability derivatives[8,23]. Across all three response branches, the SENK backbone already provided a substantially stronger baseline than DetaNet (Figure 2a-c). On QM9S, the ENK-off model achieved mean absolute errors of 0.0091 for the dipole derivatives, 0.0780 for the polarizability derivatives and 0.0267 for the Hessian, compared with 0.0182, 0.2012 and 0.0822 for DetaNet, respectively. Activating ENK further reduced these errors to 0.0081, 0.0775 and 0.0262. These values are also lower than the publicly reported EnviroDetaNet results for all three overlapping QM9S response tasks and represent the current best publicly reported performance on these benchmarks[25]. The same pattern was retained on the broader QMe14S chemical space. SENK with ENK off reached errors of 0.0117, 0.1163 and 0.0303 for the three branches, whereas the corresponding reported DetaNet values were 0.0395, 0.4673 and 0.1074. ENK activation provided an additional reduction to 0.0106, 0.1121 and 0.0279. Target-versus-prediction parity and residual distributions for all six benchmark evaluations are provided in Supplementary Fig. 6, showing close agreement with the identity relation across both datasets.

These results establish a stable global response-state baseline across both datasets. Most of the reduction in global MAE is attributable to the equivariant backbone, while ENK provides a smaller but consistent state-dependent calibration across the evaluated response branches. This separation allows the subsequent analyses to focus on how local chemical environments reshape response states and characteristic spectral features beyond what is conveyed by a global error metric. The subsequent analyses of ENK and the electronic prior are therefore not intended to compensate for a broadly inaccurate base model. Rather, they examine whether a strong geometry-based representation can maintain its overall performance while identifying the smaller subset of local response states that become less reliable as chemical complexity increases. Further branch- and distance-resolved ENK analyses are provided in Supplementary Sections 2.1-2.2 and Supplementary Fig. 1.

## 2. Full-spectrum prediction from small molecules to drug-like systems

We next tested whether the response-state accuracy in Figure 2a–c propagates to physically faithful IR and Raman spectra. The molecules in Figure 2d,e range from compact organic structures close to the small-molecule reference domain to drug-like systems with denser combinations of heteroatoms, conjugated fragments and coupled functional groups.

For ethylmalonic acid, methylsuccinic acid and N-acetylglycine, all models reproduce the principal spectral profiles, but DetaNet retains visible local frequency and intensity deviations (Figure 2d). Both SENK configurations closely follow the B3LYP/def2-TZVP references across fingerprint, carbonyl-associated and high-frequency stretching regions. The near-coincidence of the ENK-on and ENK-off spectra indicates that ENK preserves response states already represented reliably by the backbone.

The distinction becomes clearer for the drug-like systems (Figure 2e). In cefixime, DetaNet shows pronounced peak displacement and profile distortion in chemically crowded regions, whereas SENK retains the principal peak positions and intensity distribution in both IR and Raman spectra. For atorvastatin and pomalidomide, DetaNet captures the broad profiles but exhibits more localized peak shifts and intensity mismatches in regions containing coupled carbonyl, heteroatom and conjugated environments. SENK remains closer to the DFT references while preserving the overall spectral structure.

These results show that accurate tensor prediction transfers to full-spectrum behaviour across increasing molecular complexity. They also motivate a local view of generalization: a spectrum may remain accurate overall while a small number of unfamiliar response environments produce conspicuous deviations. We therefore next examine how ENK identifies reduced response-state reliability and how the electronic prior determines whether selective calibration is supported.

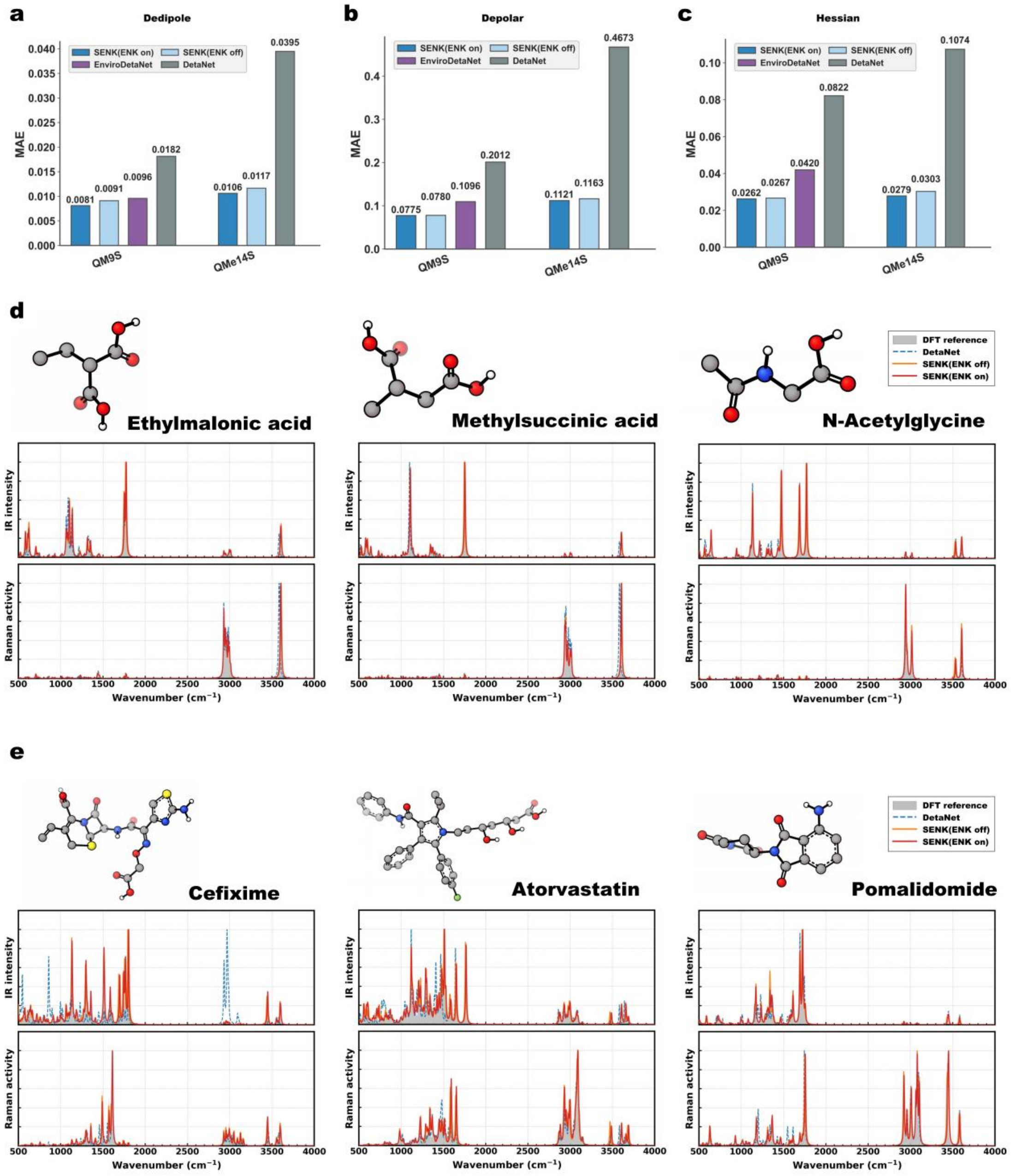

**Figure 2.** Model performance and IR/Raman spectral prediction from small molecules to drug-like systems. (a-c) MAE bar charts for dipole derivatives (Dedipole), polarizability derivatives (Depolar), and Hessian, comparing SENK (ENK on/off), EnviroDetaNet, and DetaNet on QM9S, and SENK (ENK on/off) with DetaNet on QMe14S. The EnviroDetaNet QM9S values are taken from the original EnviroDetaNet publication[25], and the DetaNet QMe14S values are taken from the original QMe14S publication[8]. (d) Spectral predictions for the QM9-like small molecules ethylmalonic acid, methylsuccinic acid, and N-acetylglycine. (e) Spectral predictions for the drug-like systems cefixime, atorvastatin, and pomalidomide. For each molecule, the molecular structure is shown above the predicted spectra, with the IR spectrum and Raman spectrum displayed in the upper and lower panels, respectively. SENK ENK-on, SENK ENK-off, and DetaNet predictions are compared with B3LYP/def2-TZVP DFT reference spectra shown as gray shaded profiles. All spectral intensities are normalized for visual comparison.

## 3. Response-state sensing and electron-informed calibration across chemical space

### 3.1 ENK identifies and refines localized response-state instability

The preceding benchmarks establish a strong geometry-to-response baseline. The role of ENK is therefore not to replace the backbone prediction, but to determine where that response representation requires state-dependent refinement. Within QMe14S, molecules with elevated response-state uncertainty, $\nu_{\mathrm{RS}}$, are sparse and concentrated in restricted regions of the molecular-size versus heteroatom-fraction plane (Figure 3a). In the external ChEMBL set, the evaluated structures extend to substantially larger molecular sizes while occupying a compositionally heterogeneous region of the same coordinate space (Figure 3b)[8,9]. At comparable molecular sizes and heteroatom fractions, $\nu_{\mathrm{RS}}$ still varies substantially, showing that these global molecular descriptors do not fully determine response-state reliability. Coordinate and aggregation definitions are given in Supplementary Sections 1.1 and 1.5.

Response-level decomposition clarifies the local role of ENK in the learned tensors. To resolve this contribution, we compare the trained SO(3) ENK mapping with an identity mapping while keeping the model state and the remaining calculations unchanged. The comparison shows that ENK preserves the shared backbone representation while providing non-uniform local refinement across response branches and geometry-defined chemical environments (Figure 3c and Supplementary Sections 2.1-2.4). Reference-anchored overlap matching further shows that these response-state changes propagate differently across spectral windows and response branches rather than producing a uniform empirical frequency or intensity shift (Supplementary Section 3 and Supplementary Fig. 2).

### 3.2 Response-state risk combines model reliability with spectral consequence

A reliability signal alone does not indicate whether an uncertain response state will materially alter the spectrum. We therefore combine the ENK-derived model-state signal with a sensitivity term that measures how strongly low-curvature, IR-active modes can amplify a response perturbation. Molecular response-state uncertainty is computed from atom-level Kalman gains:

$$\nu_{\mathrm{RS}} = 1 - K_{\mathrm{mean}} = 1 - \frac{1}{N}\sum_{i=1}^{N} K_i \,. \tag{1}$$

A larger $\nu_{\mathrm{RS}}$ indicates a lower mean ENK gain and higher model-state uncertainty. Structural softness and normalized inverse-eigenvalue mode weights are defined as:

$$s_{\mathrm{soft}} = \frac{1}{\lambda_{\mathrm{min}}^{+}} \,, \tag{2}$$

$$w_m = \frac{1/\lambda_m}{\sum_n 1/\lambda_n} \,. \tag{3}$$

and the IR-response term and combined sensitivity are:

$$s_{\mathrm{response}} = \sum_m w_m \frac{I_m}{I_{\mathrm{max}}} \,, \tag{4}$$

$$\mathcal{S} = \log(1 + s_{\mathrm{soft}}) \log(1 + s_{\mathrm{response}}). \tag{5}$$

The joint response-risk score used for case ranking is:

$$\mathcal{R} = \nu_{\mathrm{RS}} \mathcal{S} \,. \tag{6}$$

IR intensities are scaled within each molecule and the inverse-eigenvalue weights sum to one. In this construction,$\nu_{\mathrm{RS}}$ identifies response states that the learned filter regards as less reliable, whereas $\mathcal{S}$ estimates whether their perturbation is likely to produce a visible spectral consequence. Their product prioritizes the subset in which both conditions occur.

The ChEMBL analysis characterizes the joint distribution of response-state variation, spectroscopic sensitivity and EP/GSC actuation across the external chemical space. Across 4,989 structures, the calibration magnitude varies with $\nu_{\mathrm{RS}}$ and $\mathcal{S}$ (Figure 3f,g), while the magnitude of subsequent error change depends on the response branch and local chemical environment. Sample accounting and complementary chemical-space summaries are provided in Supplementary Sections 1.1 and 6 and Supplementary Fig. 5.

### 3.3 The electronic prior supplies selective, branch-specific calibration

After ENK identifies a response state requiring attention, the electronic prior determines whether a physically informed correction is available. EP consistency regularization adapts the learned representation during training, whereas runtime guided spectral calibration applies a bounded residual to eligible response branches. The two stages therefore separate representation stabilization from targeted correction.

This distinction is necessary because electronic information transfers differently across tensor tasks. In the dipole-derivative branch, the restricted DD-GSC update improves the principal deviatoric and norm-based errors beyond the intermediate EP-CR state. The representative off-diagonal Hessian branch likewise benefits from runtime GSC. For the polarizability-derivative task, the evaluated configuration retains the EP-CR representation without an additional runtime DP-GSC residual (Supplementary Section 4.5). EP/GSC therefore operates through response-specific calibration steps, yielding branch-specific electronic-prior corrections.

The local-environment analysis shows pronounced chemical-environment dependence in the representative off-diagonal Hessian branch (Figure 3d). Relative to EP-CR, the geometry-defined C-N, aromatic-like C-C, C-C and amide-like C-N strata give pooled gains of +37.4%, +19.3%, +15.2% and +11.8%, respectively. N-H, O/N-H donor, C-H and O-H are also positive at +10.2%, +7.8%, +6.3% and +4.7%, whereas high-Z element-containing, halogen-containing and carbonyl-like C-O strata give -6.5%, -23.1% and -27.9%, respectively. These overlapping geometry-defined strata reveal chemical-environment heterogeneity in the evaluated response, with formal definitions and element-pair decomposition given in Supplementary Sections 2.4 and 5.2.

The joint motif map places the ENK-derived response-state score alongside the matched EP/GSC effect for the same local strata (Figure 3e). $\nu_{\mathrm{RS}}$ is associated with larger pre-calibration error, whereas the direction and magnitude of the subsequent correction vary with the response branch and electronic environment. Together, Figure 3 establishes the division of labour within SENK: the equivariant backbone predicts the global response, ENK senses and refines chemically structured response-state variation, spectral sensitivity identifies consequential states, and EP/GSC provides bounded branch-specific correction where the electronic prior supports it.

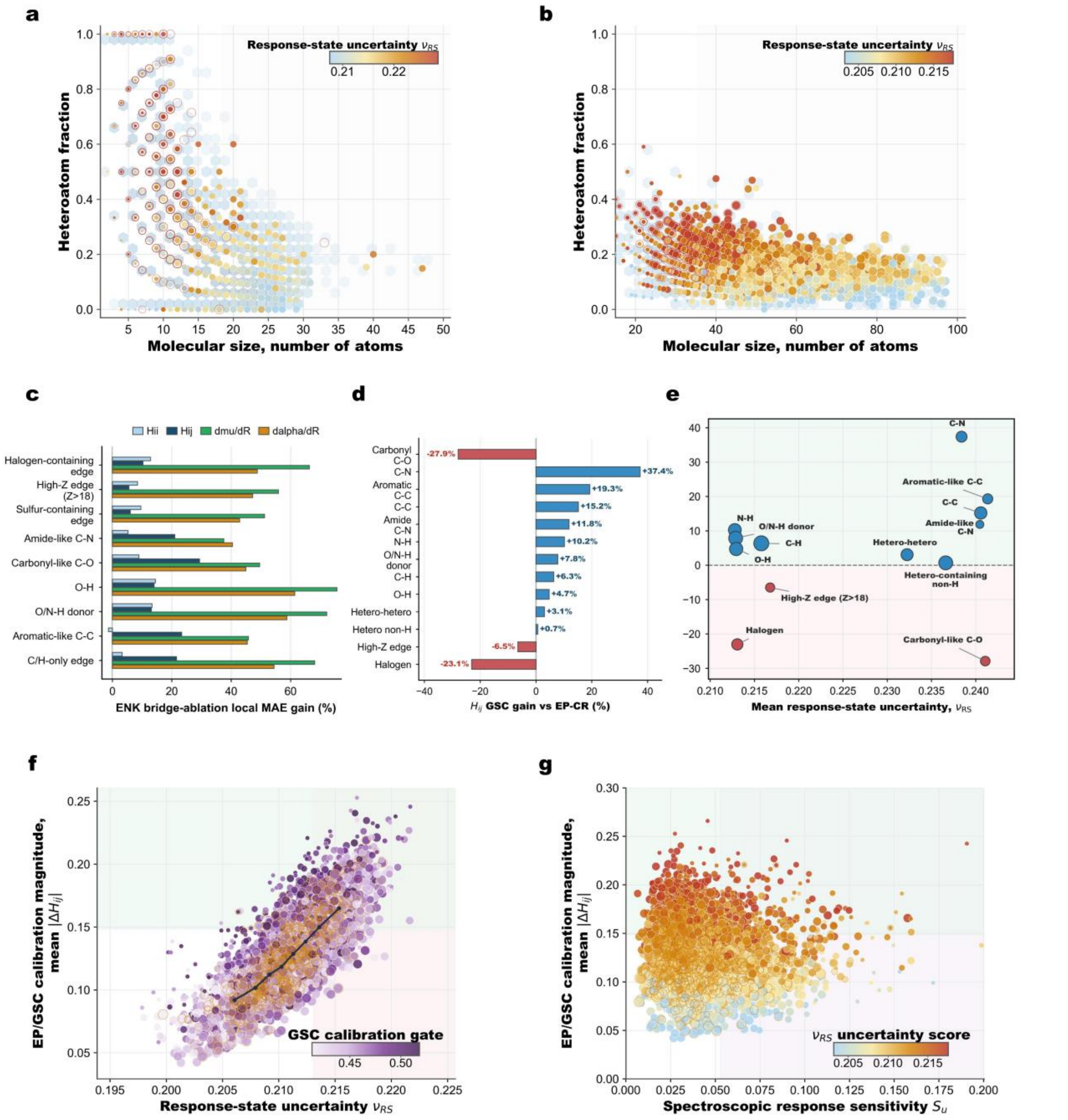

**Figure 3.** Localized response-state risk sensing and electron-informed calibration across chemical space.

(a,b) Chemical-environment maps on the molecular-size versus heteroatom-fraction plane for the QMe14S training range (a) and external ChEMBL set (b), with colour denoting the molecular ENK-derived response-state score $\nu_{RS}$; colour limits are scaled independently. (c) Local contribution of the trained SO(3) ENK mapping relative to an identity mapping under the same model state; intervals are molecule-cluster bootstrap 95% confidence intervals. (d) Matched motif-selective $H_{ij}$ MAE gains of runtime GSC relative to EP-CR. (e) Joint local map of mean $\nu_{\mathrm{RS}}$ and matched EP/GSC effect for the same geometry-defined strata. (f,g) External ChEMBL response-state and actuation analyses across 4,989 structures. Molecular counts, proxy definitions and aggregation rules are given in the Supplementary Information.

## 4. Response-state stability and electron-informed calibration in biomolecular systems

We next tested the response-state framework in biomolecular systems. For oxytocin and Leu-enkephalin, the ENK-on and ENK-off predictions remain close to the B3LYP/def2-TZVP references across the major IR and Raman regions, and the predicted IR spectral profiles are broadly consistent with the experimental IR curves (Figure 4). The agreement retained for these larger and more flexible molecules shows that increased molecular size alone does not determine spectral transferability when the relevant local response behaviour remains well described. As a separate application-oriented illustration, SENK-predicted Raman spectra can also serve as component-level references for unsaturation-sensitive lipid markers relevant to single-cell metabolic readout (Supplementary Fig. 7)[43].

The Leu-enkephalin/Met-enkephalin pair provides a controlled test of departure from this regime. The two pentapeptides share the same backbone and differ at the C-terminal residue, where methionine introduces a thioether sulfur capable of forming a Tyr-OH···Met-S contact. In the Met-enkephalin ENK-off prediction, the tyrosine O-H stretching feature is visibly blue-shifted relative to the DFT reference, whereas the corresponding region in Leu-enkephalin remains accurate. The contrast isolates a localized stereoelectronic environment rather than peptide size or backbone complexity as the source of the spectral deviation.

Enabling ENK moves the Met-enkephalin O-H feature toward the reference, indicating reduced reliability of the geometry-derived response in the uncommon O-H···S environment (Figure 5a). EP then supplies an additional calibration shift of -28 $cm^{-1}$ in both IR and Raman. Candidate-interaction screening identifies Tyr-OH···Met-S as the strongest X-H donor-acceptor contact, and NBO analysis supports an LP(S)→BD*(O-H) interaction associated with the affected bond (Figure 5b,c). The spectral changes remain concentrated in the interaction-sensitive window, while the remaining IR and Raman profiles are preserved (Figure 5d).

The 2onw:SSTSAA peptide provides an independent test in a more strongly polarized environment. Its terminal ammonium and carboxylate groups form a salt bridge, and the system was previously examined by TL-DetaNet, which required additional transfer learning on a short-peptide dataset[24]. SENK is applied here without transfer learning or system-specific fine-tuning and accurately predicts the key spectral feature for which the TL-DetaNet result showed a residual frequency offset. ENK partially shifts the ENK-off feature at 2,526 $cm^{-1}$, after which EP moves it to 2,493 $cm^{-1}$, giving a total calibration shift of -33 $cm^{-1}$ in both IR and Raman (Figure 5e,h). A related high-frequency IR feature shifts from 3,462 to 3,430 $cm^{-1}$. Interaction ranking and NBO evidence identify the terminal contact as the dominant screened local interaction (Figure 5f,g), and the calibration remains confined mainly to salt-bridge-sensitive features.

Figures 4 and 5 therefore provide complementary tests of the response-state pathway. Baseline-supported biomolecules require little intervention, whereas localized donor-acceptor or strongly polarized environments can produce spectrally consequential response-state risk. ENK provides reliability-aware refinement, and EP supplies a second, electronically informed calibration while preserving unaffected spectral regions.

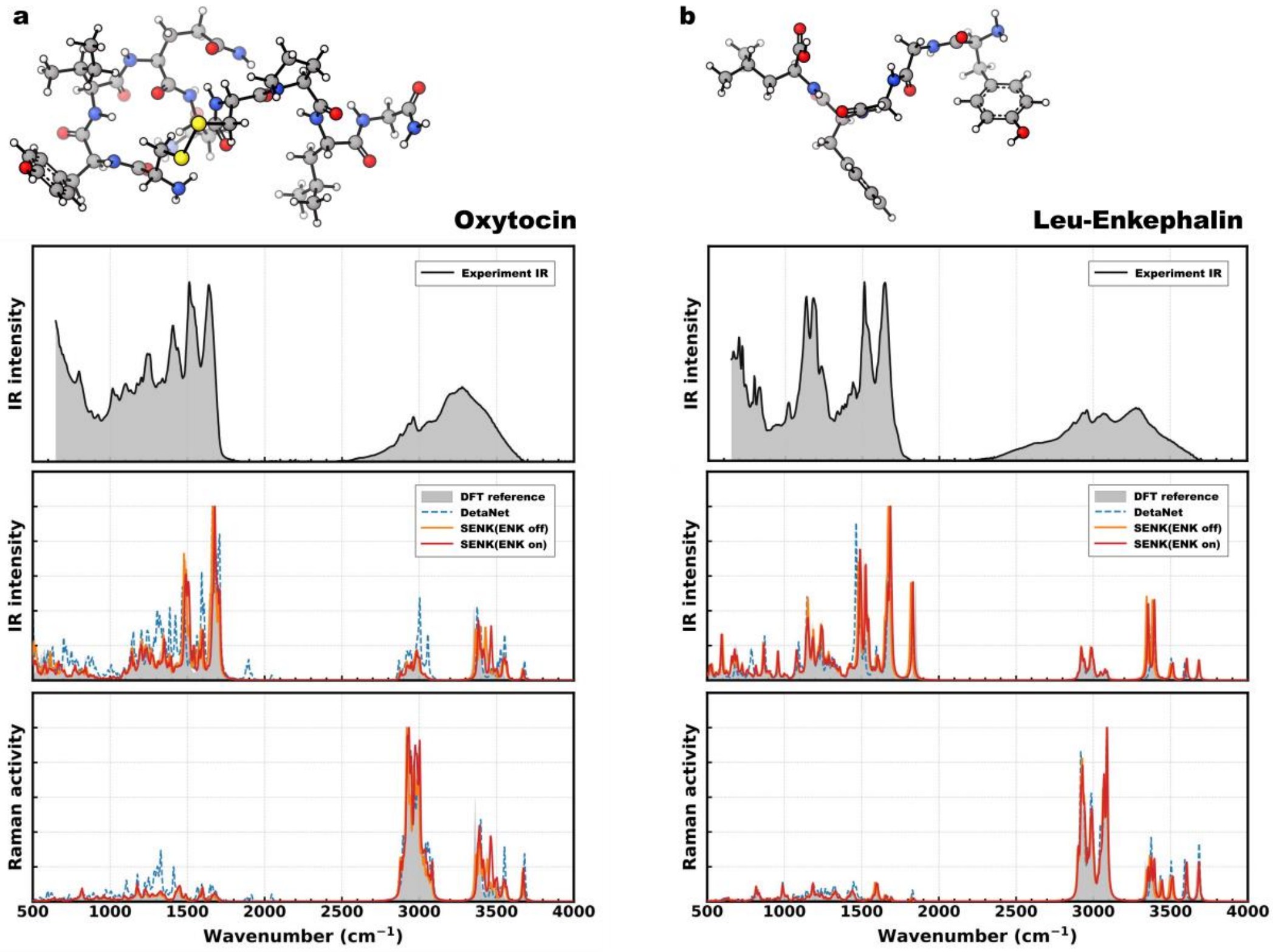

**Figure 4.** Predicted and experimental vibrational spectra of biomolecular systems. (a) Oxytocin (Cys-Tyr-Ile-Gln-Asn-Cys-Pro-Leu-Gly-NH2). (b) Leu-enkephalin (Tyr-Gly-Gly-Phe-Leu). For each system, the top panel shows the experimental IR spectrum, the middle panel shows the predicted IR spectrum, and the bottom panel shows the predicted Raman spectrum. In the predicted IR and Raman panels, SENK predictions with ENK on and ENK off are compared against the B3LYP/def2-TZVP DFT reference spectra shown as gray shaded profiles. All spectral intensities are normalized for visual comparison.

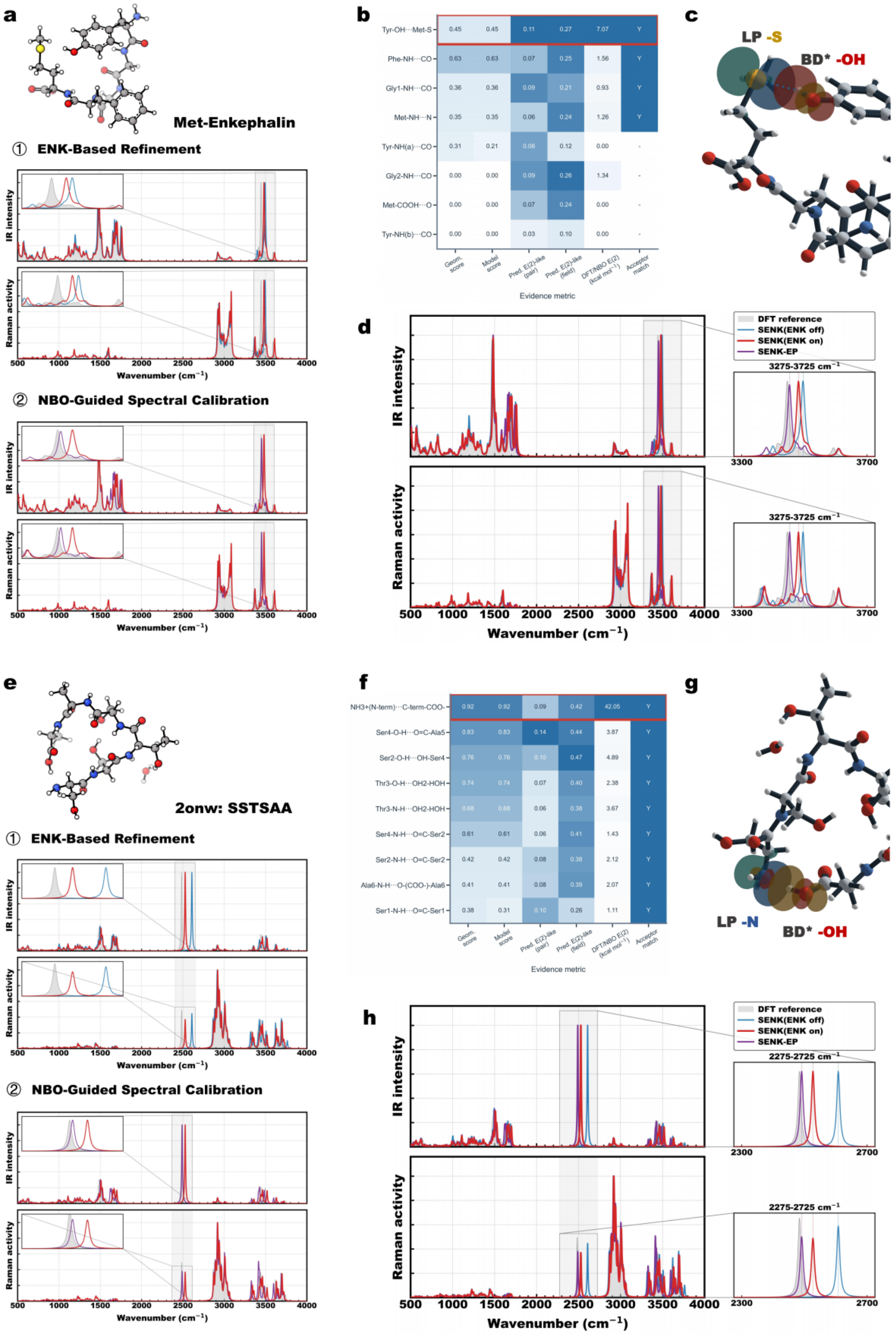

**Figure 5.** Electron-informed response-state calibration in Met-enkephalin and 2onw:SSTSAA. (a-d) Met-enkephalin. (a) Molecular structure and stepwise IR and Raman comparisons during ENK refinement and subsequent NBO-guided calibration. Insets highlight the 3350-3600 $cm^{-1}$ region. (b) Evidence matrix for model-screened X-H donor-acceptor contacts, including geometry and model scores, predicted pair- and field-level E(2)-like contributions, DFT/NBO E(2), and acceptor matching. (c) Local orbital interaction for the Tyr-OH···Met-S contact, showing LP(S) and BD*(O-H). (d) Full IR and Raman spectra with enlarged views of the 3275-3725 $cm^{-1}$ region. (e-h) 2onw:SSTSAA. (e) Molecular structure and corresponding stepwise spectral comparisons, with insets highlighting the 2400-2650 $cm^{-1}$ region. (f) Evidence matrix for the screened X-H donor-acceptor contacts. (g) Local orbital interaction associated with the terminal salt bridge, showing the donor lone pair and the corresponding X-H antibonding orbital. (h) Full IR and Raman spectra with enlarged views of the 2275-2725 $cm^{-1}$ region. Gray, DFT reference; blue, SENK with ENK off; red, SENK with ENK on; purple, SENK-EP. Orbital cube files were generated with Multiwfn[44,45].

## 5. Transferable vibrational characterization of molecular building units for organic materials

The preceding biomolecular analyses show that spectral transferability depends not only on system size, but also on the local chemical and electronic environments governing characteristic vibrational responses. We next test whether this local response-state picture extends to molecular building units relevant to organic condensed-phase and functional-material settings. We considered two complementary external tests. First, the R-3B69 benchmark probes finite, crystal-derived molecular trimers whose vibrational responses are shaped by intermolecular contacts and collective local environments[29,42]. Second, a homologous π-conjugated oligothiophene series probes whether the learned polarizability response retains an experimentally observed conjugation-dependent Raman signature[41]. Neither evaluation involves system-specific retraining or fine-tuning.

R-3B69 contains 69 crystal-derived trimers from 23 molecular families, with three geometrical configurations per family, and provides double-harmonic IR and Raman references for noncovalently interacting molecular assemblies (Figure 6a). We evaluated SENK directly on the supplied reference geometries and compared the resulting spectra with the reference spectra using the modified spectral match score, $r_{\mathrm{MSC}}=(\mathbf{u}\cdot\mathbf{v})^2/[(\mathbf{u}\cdot\mathbf{u})(\mathbf{v}\cdot\mathbf{v})]$, after independent spectral normalization and $30\mathrm{cm}^{-1}$ Lorentzian broadening. Across all 69 trimers, SENK with ENK enabled reached a mean full-spectrum Raman $r_{\mathrm{MSC}}$ of 0.836 and a median of 0.895 (Figure 6b). The corresponding mean IR value was 0.705. For context, the best published end-to-end MACE-OFF23/MACE-MDP pipeline reports mean R-3B69 values of 0.835 for Raman and 0.764 for IR[29]. The result shows that a response model trained on molecular spectroscopy can retain high full-spectrum fidelity when transferred zero-shot to finite crystal-derived molecular assemblies.

The R-3B69 decomposition further clarifies the role of ENK. Relative to the ENK-off state, ENK leaves the already strong full-spectrum Raman agreement essentially unchanged, with a family-averaged change of $\Delta r_{\mathrm{MSC}} = +0.003$, but improves the 400–1800$\mathrm{cm}^{-1}$ fingerprint region by $\Delta r_{\mathrm{MSC}} = +0.036$. The fingerprint improvement is observed in 82.6% of the 23 molecular families, with a family-bootstrap 95% confidence interval of [+0.019, +0.056]. The corresponding IR fingerprint change is $\Delta r_{\mathrm{MSC}} = +0.056$, with improvement in 78.3% of the molecular families and a 95% confidence interval of [+0.028, +0.090]. Representative acetamide-trimer spectra illustrate that the predicted peak pattern is retained across both fingerprint and high-frequency regions while local peak positions and intensities remain close to the reference (Figure 6c,d). These results are consistent with the response-state interpretation developed above: ENK does not impose a uniform spectral correction, but can refine selected response environments while preserving an already transferable global prediction.

We then tested a different materials-relevant limit in which the target observable is not intermolecular spectral fidelity but the evolution of a characteristic Raman response with molecular conjugation. SENK-EP was applied without retraining to a homologous 4T–10T oligothiophene series measured experimentally in dilute solution (Figure 7). We focused on the in-phase C=C, R-like vibrational subspace, for which the experimental differential Raman cross-section increases strongly with conjugation length. After normalization to 4T, the experimental responses are 1.00, 3.49, 6.46 and 8.27 for 4T, 6T, 8T and 10T, respectively, whereas SENK-EP gives 1.00, 2.84, 4.13 and 8.33. The predicted response therefore preserves the experimental ordering across all four homologues and shows a strong trend-level correlation $r = 0.938$, $n = 4$; notably, the predicted 10T/4T response ratio of 8.33 closely matches the experimental value of 8.27.

Together, the R-3B69 and oligothiophene tests extend the local response-state picture across two distinct forms of chemical-environment variation. R-3B69 probes vibrational responses shaped by explicit intermolecular contacts in crystal-derived molecular assemblies, whereas the oligothiophene series probes the evolution of polarizability response with π-conjugation. Their transfer without system-specific retraining connects local response-state learning to molecular building units relevant to organic condensed-phase and functional-material spectroscopy.

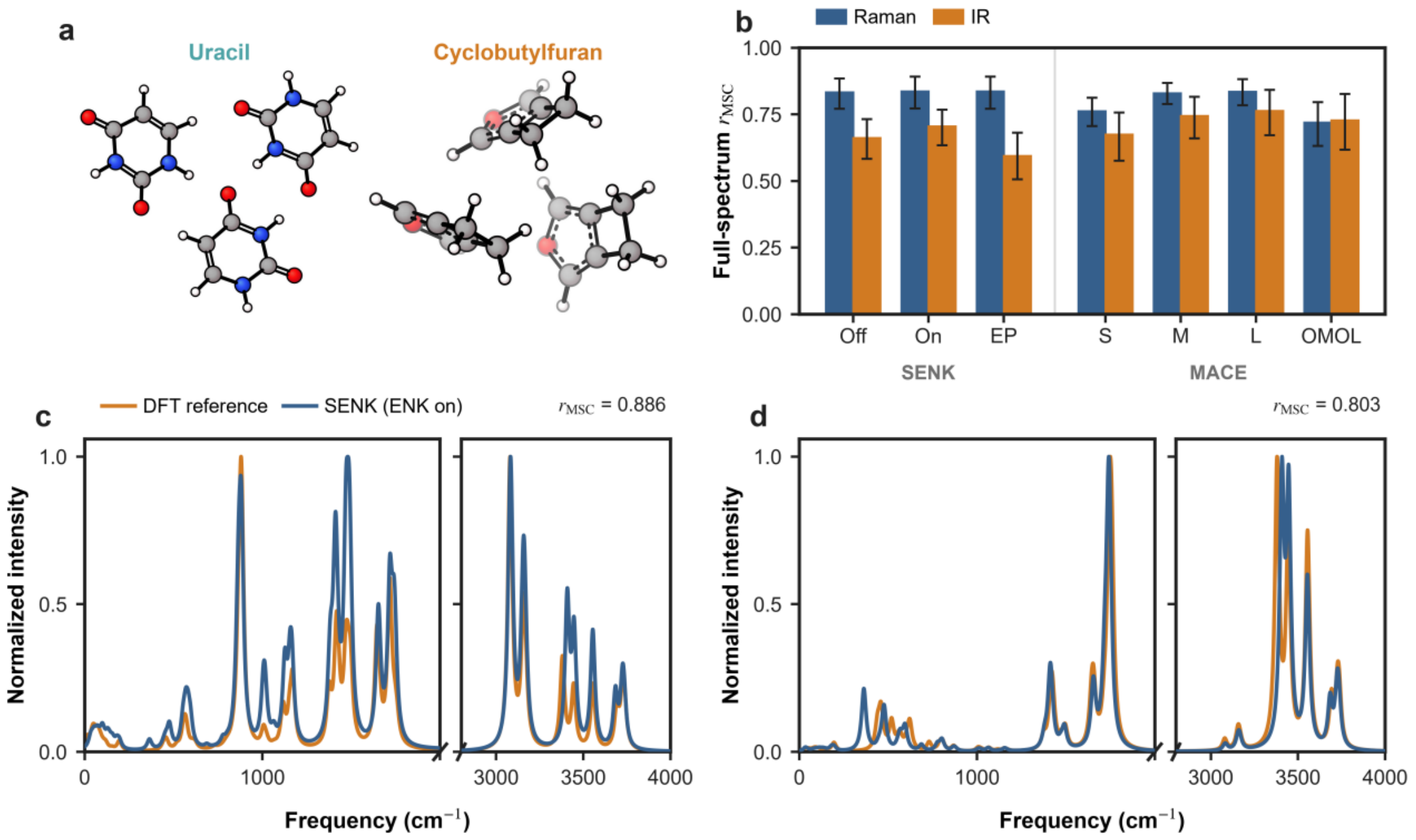

**Figure 6.** Vibrational-spectrum transfer to crystal-derived molecular assemblies in R-3B69. (a) Representative noncovalently interacting trimers. (b) Full-spectrum ($r_{\mathrm{MSC}}$) for SENK variants and published MACE-MDP baselines across 69 trimers. (c,d) Representative Raman (c) and IR (d) spectra for the 09a-acetamide trimer, comparing the DFT reference with SENK (ENK on).

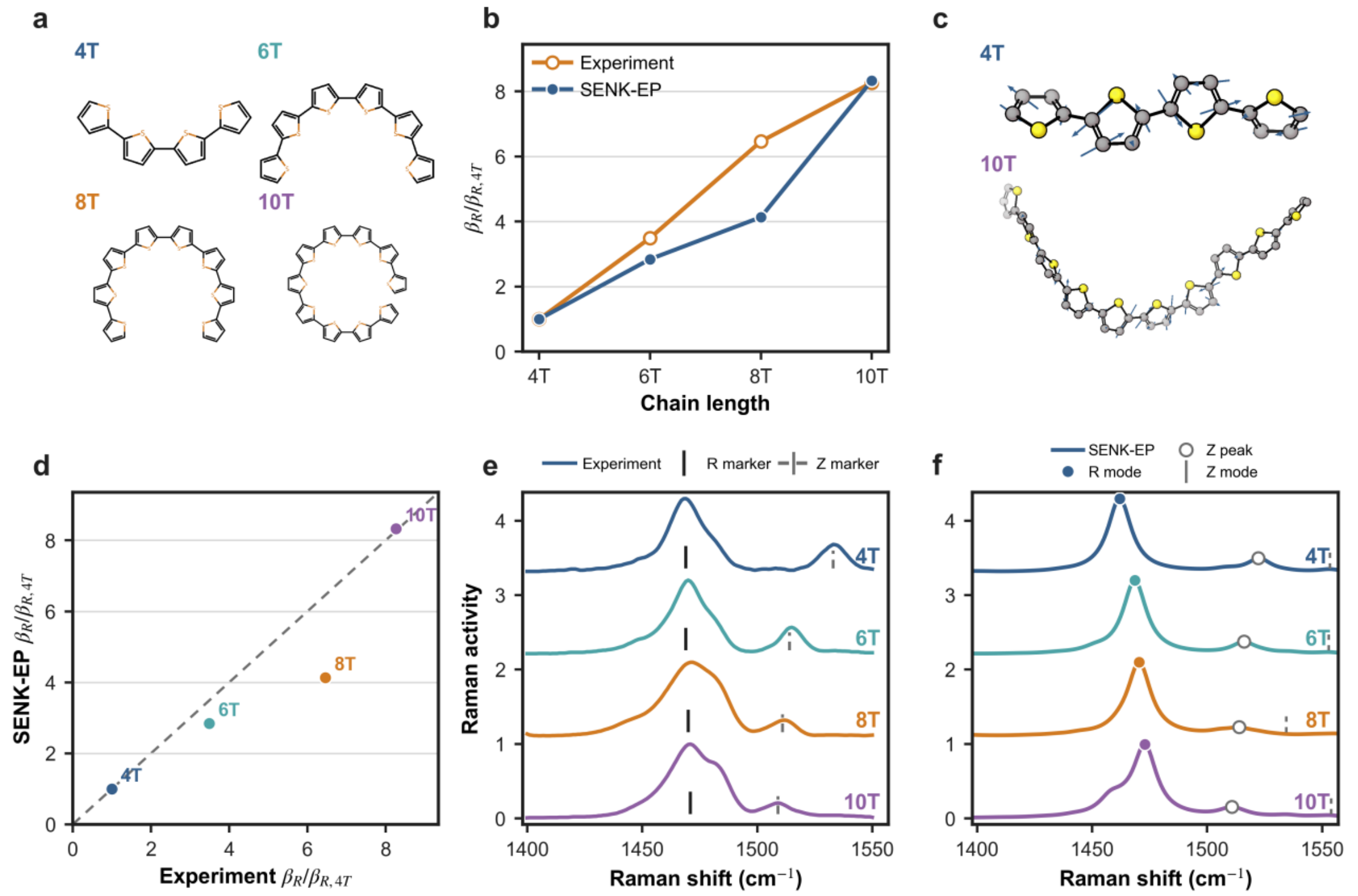

**Figure 7.** Raman characterization of π-conjugated oligothiophene semiconductor building blocks. (a) Molecular structures of the 4T, 6T, 8T and 10T series. (b–d) Experimental and SENK-EP relative R-like Raman responses, representative vibrational displacements and molecule-level comparison. (e,f) Experimental and SENK-EP Raman spectra across the oligothiophene series.

## Discussion

Chemically characteristic vibrational signals have a local physical origin: functional-group responses are modulated by their chemical and electronic environments, so cross-system spectral generalization cannot be reduced to molecular size alone. This motivates treating the Hessian, dipole derivatives and polarizability derivatives as intermediate response states through which local environmental changes are connected to spectral observables. SENK implements this response-state perspective by combining an equivariant global response model with learned reliability sensing and branch-specific, NBO-informed electronic calibration.

Across QM9S and QMe14S, the equivariant backbone provides a strong response-state baseline, with ENK yielding consistent additional improvements across the evaluated response branches. The same response-state framework extends to larger drug-like molecules and biomolecular systems. In Met-enkephalin and 2onw:SSTSAA, localized sulfur-associated and salt-bridge environments are associated with spectrally consequential response deviations, and the sequential

ENK and EP corrections shift the affected IR and Raman features toward the reference while leaving the broader spectral profiles largely unchanged. Together with the NBO analysis, these cases provide an interpretable electronic-structure context for the observed localized stereoelectronic effects[38–40].

The selective behavior of ENK and EP/GSC also provides a practical basis for extending the framework. Systematic characterization of broader chemical spaces can guide targeted expansion of response-state coverage across element combinations, functional motifs, charge states, bonding patterns and non-covalent interactions. The R-3B69 results add an external test at the level of crystal-derived molecular assemblies: SENK preserves high full-spectrum Raman similarity across 69 noncovalently interacting trimers, while ENK produces reproducible family-level refinement in the fingerprint region. This behavior complements the oligothiophene result, in which the learned polarizability response preserves an experimental conjugation-dependent Raman trend across semiconductor building blocks. Together, these tests suggest that transferable response states can remain informative across both intermolecularly organized molecular units and conjugated functional molecules.

SENK provides a coherent route from response-state learning to localized reliability sensing and electron-informed calibration for vibrational spectroscopy. The framework maintains accurate response prediction across diverse molecular classes while selectively refining spectrally consequential deviations associated with local stereoelectronic environments. Its performance in drug-like and biomolecular systems shows that this response-state representation remains effective as molecular size and chemical complexity increase. Beyond these molecular evaluations, the R-3B69 benchmark and oligothiophene Raman analysis further show that the learned structure-response relationships can transfer to crystal-derived molecular assemblies and π-conjugated semiconductor building blocks without system-specific retraining. Together, these results support transferable vibrational spectroscopic characterization across molecular and materials-relevant chemical space.

## Methods

### 1. Reference Data

QM9S and QMe14S provide direct supervision for SENK spectroscopic prediction heads[8,23],

including Hessian matrix blocks and dipole moment and polarizability derivatives predicted from equilibrium geometries. QM9S serves as the reference small-molecule spectroscopy dataset, whereas QMe14S serves as the main target dataset. SIMG and qcMol are used to train the NBO-based electron prior: SIMG provides heterogeneous NBO graph supervision for transferable electronic pretraining, and qcMol provides atom- and bond-resolved quantum-chemical annotations for joint adaptation and broadened coverage[46,47]. The electron-prior training corpus combines a PubChemQC/ZINC-derived molecular subset with a PDBbind 2020 ligand subset to cover both broad drug-like chemistry and recognition-relevant ligand environments. Reference IR and Raman spectra were calculated using Gaussian 16 software[48]. Geometry optimizations and harmonic frequency calculations were performed at the B3LYP/def2-TZVP level of theory, and the IR intensities and Raman activities were obtained from the corresponding frequency calculations[49,50]. Experimental IR spectra were recorded using a Thermo Scientific Nicolet iS50 Fourier-transform infrared spectrometer over the wavenumber range of 4000-500 $cm^{-1}$. Molecular structures shown in the figures were rendered using XYZRender based on the optimized geometries[51].

**2. SENK architecture**

The SENK model predicts four spectroscopically relevant per-atom and per-edge quantities—the diagonal Hessian matrix block $H_{ii}$, the off-diagonal Hessian matrix block $H_{ij}$, the dipole derivative $\partial\mu/\partial R$, and the polarizability derivative $\partial\alpha/\partial R$—directly from a molecular equilibrium geometry. These four predicted quantities are subsequently passed to a harmonic normal-mode analysis pipeline to generate infrared (IR) absorption and Raman scattering spectra. Formally, given atomic numbers $\{z_i\}$ and equilibrium Cartesian coordinates $\{r_i\}$, the model constructs a spatial radius graph. The core representation engine relies on an SO(3)-equivariant EquiformerV2 architecture[52]. Two complementary modules are attached to this backbone. ENK performs an environment-dependent confidence update on selected equivariant channels before tensor readout. EP supplies geometry-derived electronic descriptors at two levels: a trainable equivariant injection conditions the latent representation, and a bounded post-forward calibration can act on chemically localized residuals. The resulting architecture is organized into three coupled components: robust geometric representation learning, representation-level environmental adaptation, and electronic-structure-guided tensor correction, all built on a common equivariant backbone.

## 3. EquiformerV2 Adaptation for Spectroscopic Prediction

The standard EquiformerV2 architecture was adapted to spectroscopic regression by removing the Open Catalyst Project dependency and replacing it with a lightweight radius-graph constructor tailored to the QMe14S and QM9S spatial scales[52]. The Hessian blocks are computed by two derivative routes that share the same equivariant representation but keep the numerical structure of $H_{ii}$ and $H_{ij}$ separate. For $H_{ii}$ we use a two-leaf mixed-derivative construction to isolate the second-derivative graph cleanly. For $H_{ij}$ we retain the edge-wise derivative route, which follows the interacting atom pair directly and preserves the off-diagonal curvature signal. For the main experiments, SENK used 5.0 Å radius graphs and an EquiformerV2 backbone with $l_{\max} = 3$, $m_{\max} = 2$, 128 hidden channels and 128 radial basis functions. Models were trained using an 80:10:10 train/validation/test split and AdamW optimization with a batch size of 32, an initial learning rate of $5 \times 10^{-4}$ and a weight decay of $1 \times 10^{-5}$. Training was performed for up to 200 epochs with cosine learning-rate decay to $1 \times 10^{-6}$ and gradient clipping at 3.0.

For the supervised branches, the reported main experiments use a unified mean absolute error objective. For branch $t$ with prediction $\hat{y}^{(t)}$ and target $y^{(t)}$, we minimize

$$\mathcal{L}^{(t)} = \frac{1}{N_t} \sum_{n=1}^{N_t} \left\| \hat{y}_n^{(t)} - y_n^{(t)} \right\|_1 \tag{7}$$

with tensor outputs compared element-wise in their native shape.

The dipole branch uses a complementary equivariant vector readout: a scalar coefficient from the invariant channel multiplies the center-of-mass-relative coordinate, and a projected $l = 1$ feature path supplies an additive vector term. The dipole derivative is then obtained by differentiating the molecular dipole with respect to Cartesian coordinates.

The polarizability branch uses the native rank-2 equivariant readout of the EquiformerV2 backbone rather than collapsing all higher-order information into scalar invariants as in Equiformer[53]. Additional design choices improve numerical stability on derivative tasks: multi-layer SO(3)/SO(2) transformer blocks with per-block spherical-harmonic normalization, edge-degree embeddings, atom-edge embeddings, label-magnitude ceilings for outlier filtering, non-finite value filtering, gradient clipping, and position centering. The atomic embedding table is initialized over a broad element range to avoid inference-time patching for unseen elements. These implementation choices make the spectroscopic adaptation of EquiformerV2 numerically stable and compatible with chemically diverse molecules.

## 4. Equivariant Neural Kalman (ENK) Filtering

Drawing conceptual inspiration from dynamic recursive state estimation in object tracking[54–56], we designed ENK to separate stable, chemically meaningful feature signals from local geometric perturbations at the intermediate SO(3) representation level. ENK is implemented as a learned, equivariance-preserving confidence filter on the SO(3) latent representation. Filtering is applied only to the $l = 0$ and $l = 2$ readout channels, which are the linear readout paths routed through ENK in the implemented architecture. For each filtered degree, the state update, process uncertainty, and observation-noise logit are predicted from the invariant features $x^0$:

$$x^l_{\text{pred}} = x^l \odot \left[1 + \tanh\left(f^l_{\text{state}}(x^0)\right)\right], \tag{8}$$

$$q^l = \text{softplus}\left(f^l_Q(x^0)\right), \tag{9}$$

$$\tilde{r}^l = f^l_R(x^0). \tag{10}$$

When an atom-level electron prior $\pi_i$ is available, it modulates the observation-noise logit through a zero-initialized FiLM transformation:

$$\tilde{r}^l \leftarrow \tilde{r}^l[1 + \tanh(W^l_s \pi_i)] + \tanh(W^l_t \pi_i), \tag{11}$$

$$r^l = \text{softplus}(\tilde{r}^l). \tag{12}$$

The filtered state is obtained with an atom-resolved Kalman gain:

$$K^l = \frac{q^l}{q^l + r^l + \varepsilon}, \tag{13}$$

$$x^l_{\text{filt}} = x^l_{\text{pred}} + K^l\left(x^l - x^l_{\text{pred}}\right). \tag{14}$$

$K^l$ is a scalar for each atom and filtered degree and is broadcast over the magnetic components, thereby preserving rotational equivariance. A small $K^l$ means that the local backbone observation is assigned low confidence and the filtered state remains closer to the learned state prior, whereas a large $K^l$ preserves the observation. The bounded multiplicative state update and the restriction to the linear $l = 0$ and $l = 2$ paths avoid uncontrolled amplification through the quadratic Clebsch-Gordan self-couplings used by other readout channels. ENK should therefore be interpreted as a learned confidence-weighted representation update rather than a generic low-pass denoiser. Because this update is inserted before the final tensor readout, the first correction occurs on the intermediate equivariant representation itself rather than on a post-hoc output tensor. In the ENK-EP cascade, ENK acts first in the intermediate SO(3) feature space by judging whether a local chemical environment is confidently represented or needs partial correction; EP then uses this judgment together with the NBO prior to apply a selective, bounded second-stage correction only where the electronic proxy is informative.

## 5. Multi-Stage Electron Prior (EP)

### 5.1 NBO Predictor and Multi-Stage Transfer from SIMG to qcMol

Instead of adding a standalone electronic-structure engine in parallel with SENK, we use a geometry-driven NBO predictor based on EMPP design principles[57]. The predictor separates generic geometric encoding from explicit electronic supervision: atomic geometry is first encoded into atom features, then expanded into a heterogeneous graph with atom, bond, and lone-pair-related tokens, and finally refined by equivariant message passing and pair-aware aggregation. Decoupled output heads allow different electronic observables to be learned without forcing them into a single regression target.

SIMG and qcMol provide complementary supervision for this shared electronic representation[46,47]. In SIMG, the predictor is trained on heterogeneous NBO graph labels consisting of atom-level descriptors, bond-level descriptors, interaction-level descriptors, and binary topology links. This stage teaches the model to recover a transferable electronic scaffold from geometry, including charge redistribution, bonded orbital character, and donor-acceptor tendencies. The qcMol stage refines the same representation with atom- and bond-resolved quantum-chemical observables. The default atom targets are NAO, LP, and NPA; the default bond target is NBO; and auxiliary targets include ADCH, LI, and ELF on atoms together with DI, LBO, and Mayer quantities on bonds. In other words, qcMol does not replace the SIMG formulation; it anchors the same backbone to a richer and more interpretable quantum-chemical target space.

Training follows a three-stage transfer curriculum. Stage 1 pretrains the EMPP-based predictor on SIMG. Stage 2 jointly optimizes SIMG and qcMol so that transferable electronic features are retained while the model aligns to qcMol observables. Stage 3 shifts to a qcMol-dominant regime with a small amount of SIMG replay to reduce forgetting. The qcMol component also broadens molecular coverage beyond the original SIMG distribution. The PubChemQC/ZINC subset contributes broad drug-like molecule chemistry, whereas the PDBbind subset adds protein-ligand examples biased toward bioactive bound ligands and recognition-relevant chemistry. The resulting electron prior is therefore exposed to a wider range of chemically realistic bonding motifs and donor-acceptor situations.

### 5.2 Construction and Injection of the Electron Prior

At inference time, the NBO predictor serves as a geometry-conditioned surrogate model for local electronic descriptors. Given only the equilibrium coordinates and atomic numbers, it predicts atom-level and bond-level NBO quantities such as natural atomic charges, NAO populations, and bond occupancies. These predictions are denormalized and mapped through learned adapters into two complementary prior fields. The atom prior $\pi_{\text{atom}}$ summarizes local population and valence-state cues for each atom. The edge prior $\pi_{\text{edge}}$ encodes pairwise chemical information, with bonded edges built from predicted bond descriptors and nonbonded edges synthesized from atom-level predictions and pairwise geometry. Long-range donor-acceptor interactions are folded back onto the corresponding atoms and edges so that resonance and charge-transfer tendencies can influence the spatial graph even between nonbonded sites.

Rather than being concatenated only at the final readout, the prior is converted into additive SO(3)-equivariant tensor corrections and injected into deeper EquiformerV2 blocks at an internal position that we refer to as Point A:

$$X^{(b+1/2)} \leftarrow X^{(b+1/2)} + \Delta_{\text{EP}}^{(b)}(\pi_{\text{atom}}, \pi_{\text{edge}}) \,. \tag{15}$$

This placement is deliberate: geometric context has already been established through attention, but task-specific readout specialization has not yet occurred. The injected electronic correction can therefore reshape the intermediate representation in a chemically informed way. The prior-conditioned features then pass to ENK and the spectroscopic heads, where the atom prior further modulates the ENK observation-noise estimate through FiLM conditioning. Zero-initialized injection projections make the initial Point-A perturbation vanish, while warm-up-controlled atom, edge, and injector gates introduce the electronic pathway progressively during EP adaptation. In the actual implementation, EP therefore plays three coupled roles: it reshapes backbone features, informs ENK filtering, and anchors the final spectroscopic readouts in a chemically grounded intermediate representation.

### 5.3 NBO Consistency-Regularized EP Adaptation

After pretraining the NBO predictor, the electronic prior is adapted to each spectroscopic branch through NBO consistency regularization (CR). During this stage, the inherited SENK backbone, ENK module, task readout, and electron-prior predictor are held fixed; the trainable parameters are concentrated in the SO(3) prior injector, the atom- and edge-prior adaptation branch and gates, and the small branch-specific consistency model. The EP objective is

$$\mathcal{L}_{\mathrm{EP}}^{(t)} = \mathcal{L}_{\mathrm{sup}}^{(t)} + \lambda_{\mathrm{CR}}(e)\mathcal{L}_{\mathrm{CR}}^{(t)}, \tag{16}$$

where $\lambda_{\mathrm{CR}}(e)$ and the injection gates are increased gradually during a short warm-up. The NBO predictions used as consistency targets are detached, so CR adapts the spectroscopic response to the electronic prior without retraining the NBO predictor.

The consistency relation is selected according to the physical role of each tensor. For a bonded edge $e = (i, j)$ with unit direction $\hat{R}_e$, the longitudinal Hessian coupling is

$$h_e = \left|\hat{R}_e^{\mathrm{T}} \hat{H}_{ij,e} \hat{R}_e\right|. \tag{17}$$

The $H_{ij}$ consistency loss imposes a consensus-weighted Badger-type relation between $h_e$ and the predicted NBO bond occupancy $\omega_e$:

$$\mathcal{L}_{\mathrm{CR}}^{H_{ij}} = \frac{1}{|\mathcal{E}_{\mathrm{b}}|} \sum_{e \in \mathcal{E}_{\mathrm{b}}} s_e \left[\log h_e - (\log k + \beta \log \omega_e)\right]^2, \tag{18}$$

Here $s_e$ is a detached consensus score obtained by comparing independent bond-order indicators, so uncertain electronic estimates contribute less strongly to training. For the dipole derivative tensor $Z_i^* = \partial\mu/\partial r_i$, CR uses the translational charge-sum constraint. For a batch of $B$ molecules,

$$\mathcal{L}_{\mathrm{CR}}^{\mathrm{DD}} = \frac{1}{9B} \sum_{b=1}^{B} \left\| \frac{\sum_{i \in b} Z_i^* - Q_{\mathrm{NPA},b}\mathbf{I}}{\sqrt{N_b}} \right\|_F^2, \tag{19}$$

where $Q_{\mathrm{NPA},b}$ is the molecular charge obtained by summing the predicted NPA charges. For the polarizability derivative $T_i^{\mathrm{DP}} \in \mathbb{R}^{3\times6}$, an atom-resolved delocalization proxy $d_i$ is constructed from the predicted atomic population and incident bond occupancies. The response magnitude used by the implementation is the mean row norm:

$$m_i^{\mathrm{DP}} = \frac{1}{3}\sum_{a=1}^{3} \left\| T_{i,a,:}^{\mathrm{DP}} \right\|_2 , \tag{20}$$

which is regularized in log space:

$$\mathcal{L}_{\mathrm{CR}}^{\mathrm{DP}} = \frac{1}{N}\sum_{i} \left[\log m_i^{\mathrm{DP}} - (\log c + \delta \log d_i)\right]^2 . \tag{21}$$

These branch-specific constraints are the core of EP-CR: the electronic prior is not converted into one universal correction target, but is coupled to each spectroscopic tensor through the physical relation appropriate to that tensor.

### 5.4 NBO-Guided Spectral Calibration

NBO-guided spectral calibration (GSC) is the second EP stage. It operates after the CR-adapted SENK forward pass and leaves all learned weights unchanged. The central motivation is that vibrational-spectral errors are often localized: a molecule may contain only a few chemically unusual or interaction-sensitive bonds whose local electronic environment is not confidently represented, while most bonds are already predicted adequately. GSC therefore avoids global rescaling and instead activates bounded local corrections only where independent signals agree that a specific site is both electronically mischaracterized and spectroscopically relevant.

The principal GSC formulation acts on $H_{ij}$ because this off-diagonal Hessian block directly controls interatomic force coupling and hence the normal-mode frequencies. For a bonded edge, let the signed longitudinal coupling be $\ell_e = \hat{R}_e^{\mathrm{T}} \hat{H}_{ij,e} \hat{R}_e$ and its magnitude be $h_e = |\ell_e|$. The magnitude is compared with the NBO occupancy in a bond-class-resolved, training-set-normalized space:

$$z_e^H = \frac{\log h_e - \mu_{c(e)}^H}{\sigma_{c(e)}^H} , \tag{22}$$

$$z_e^\omega = \frac{\log \omega_e - \mu_{c(e)}^\omega}{\sigma_{c(e)}^\omega} . \tag{23}$$

A bounded electronic residual ratio is then constructed:

$$\rho_e = \operatorname{clip}(1 + \beta z_e^\omega - z_e^H, 1 - \eta, 1 + \eta) . \tag{24}$$

Rather than relying on a predefined motif list, the correction is activated by a task-specific policy score that fuses three classes of evidence. First, the physical residual measures whether the predicted longitudinal coupling is inconsistent with the NBO-implied bonded interaction. Second, the NBO branch contributes a reliability estimate based on the agreement among independent bond-order indicators together with the available bond-class support in the training set. Third, the local-environment term determines whether the edge lies in a chemically unusual regime. This is where ENK and GSC couple most directly: ENK provides an atom-resolved representation-confidence signal through its Kalman gains and branch-disagreement statistics, and GSC augments that signal with bond-class novelty, unseen local environment signatures, hydrogen-bond geometry, and donor-acceptor interaction evidence. A mode-aware gate further restricts calibration to bonds that actually participate in the relevant normal modes, so the correction is directed toward local tensor components that can shift vibrational peaks rather than uniformly perturbing the entire spectrum.

Denoting the resulting bounded edge score by $g_e$, the calibrated $H_{ij}$ update is applied only along the bond-longitudinal projector:

$$\widehat{H}_{ij,e}^{\mathrm{GSC}} = \widehat{H}_{ij,e} + \alpha_{H_{ij}}(\rho_e - 1) g_e \ell_e \big(\widehat{R}_e \otimes \widehat{R}_e\big)\,. \tag{25}$$

The corresponding diagonal blocks are updated by a $\gamma_{\mathrm{comp}}$-weighted sum of the incident $H_{ij}$ increments, providing controlled compensation toward acoustic-sum-rule consistency:

$$\widehat{H}_{ii}^{\mathrm{GSC}} = \widehat{H}_{ii} - \gamma_{\mathrm{comp}} \sum_{e:\ \mathrm{dst}(e)=i} \Delta H_{ij,e}\,. \tag{26}$$

This projector form leaves transverse bending and torsional couplings unchanged while allowing targeted adjustments to the bond-stretching component that most directly controls peak positions. In this sense, GSC is not a generic post-hoc rescaling layer, but a selective local second-stage correction that uses ENK to judge where the backbone is uncertain and uses the electron prior to determine how that specific site should be corrected.

The same calibration interface can be instantiated for intensity-response tensors, but the correction laws remain task specific. Dedipole uses a bounded bond-directional, trace-free refinement together with the charge-sum constraint, whereas the evaluated depolar calibration candidate uses a delocalization-conditioned multiplicative form. The reported task-specific configuration applies runtime GSC to $H_{ij}$, permits a restricted dedipole refinement and retains the EP-CR representation for the polarizability-derivative branch without an additional runtime Depolar-GSC residual. The matched runtime audit is reported in Supplementary Section 4.5.

## 6. Evaluation and local-response analyses

We calculated MAE and RMSE for response quantities, invariants, frequencies and peaks as

$$\mathrm{MAE} = \frac{1}{N}\sum_{i=1}^{N} |y_i - \hat{y}_i| , \tag{27}$$

$$\mathrm{RMSE} = \left[\frac{1}{N}\sum_{i=1}^{N} (y_i - \hat{y}_i)^2\right]^{1/2} . \tag{28}$$

Percentage gain relative to the stated baseline was defined as

$$G(\%) = 100 \times \frac{E_{\mathrm{baseline}} - E_{\mathrm{method}}}{E_{\mathrm{baseline}}}, \tag{29}$$

where positive gain denotes lower error. Main Fig. 3a,b use molecular response-state uncertainty

$$\nu_{\mathrm{RS}} = 1 - \bar{K} = 1 - \frac{1}{N_{\mathrm{atom}}}\sum_{i=1}^{N_{\mathrm{atom}}} K_i , \tag{30}$$

where $K_i$ denotes the atom-level Kalman gain after averaging over the ENK-enabled SO(3) degrees used in the corresponding evaluation. Molecular size is the total number of atoms including H, and the heteroatom fraction is

$$f_{\mathrm{hetero}} = \frac{N_{\mathrm{non-C,H}}}{N_{\mathrm{all\ atoms}}}. \tag{31}$$

Edges were classified as covalent-like when

$$r_{ij} \le 1.25\left(r_i^{\mathrm{cov}} + r_j^{\mathrm{cov}}\right) + 0.15\ \text{Å}. \tag{32}$$

Local motifs are geometry-defined, overlapping descriptive strata rather than mutually exclusive functional-group assignments. Proxies are: carbonyl-like C-O, covalent C-O≤1.36 Å; amide-like C-N, covalent C-N≤1.65 Å with a covalent C-O≤1.36 Å neighbour at C; aromatic-like C-C, C-C 1.20-1.50 Å with each C having at least two carbon neighbours within 1.55 Å; O/N-H donor, O-H or N-H; halogen, an F/Cl/Br endpoint; and high-Z, an endpoint with Z>18.

Mode analyses used target/DFT modes as reference anchors; ENK-on and ENK-off modes were matched independently by maximum absolute normalized vector overlap, retaining records only when both overlaps were at least 0.5. Stretching, bending and torsional labels used displacement-derived internal-coordinate scores, and peak records were selected from reference intensities.

For Main Fig. 3c, the trained ENK mapping was compared with an identity mapping while holding the model state and all remaining calculations fixed. Molecules defined the resampling clusters: error sums and counts were first aggregated within each molecule and motif. We generated B=2,000 bootstrap replicates by sampling molecules with replacement and recomputed the pooled local MAEs and percentage gain for each replicate. The 95% confidence intervals were defined by the 2.5th and 97.5th percentiles of the bootstrap distribution. Detailed internal-coordinate equations, distance bins, peak windows and chemical-space binning procedures are provided in Supplementary Sections 2.2-2.4, 3.1 and 6.2-6.5.

## 7. Materials-relevant molecular building-unit evaluation

### 7.1 R-3B69 noncovalent molecular-assembly benchmark

R-3B69 was used to evaluate transfer of SENK to noncovalently interacting molecular assemblies. The benchmark contains 69 crystal-derived trimers from 23 molecular families, with three configurations per family and double-harmonic IR and Raman reference spectra calculated at the ωB97M-D3(BJ)/def2-TZVPPD level. SENK was evaluated directly on the released geometries without benchmark-specific retraining or fine-tuning. ENK-off and ENK-on predictions were generated using the same trained model configuration. Because the stored SENK frequencies include the predefined global factor of 0.965, predicted frequencies were divided by 0.965 to recover the unscaled harmonic-frequency convention used by R-3B69.

Reference and predicted stick spectra were independently normalized and broadened with a Lorentzian function of 30 $\mathrm{cm^{-1}}$ full width at half maximum on a 0-4000 $\mathrm{cm^{-1}}$ grid. Spectral agreement was quantified using the match score $r_{\mathrm{MSC}} = (\mathbf{u}\cdot\mathbf{v})^2/[(\mathbf{u}\cdot\mathbf{u})(\mathbf{v}\cdot\mathbf{v})]$. In addition to the full spectral range, 0-400, 400-1800 and 2800-4000 $\mathrm{cm^{-1}}$ windows were analyzed separately. Paired ENK effects were averaged over the three configurations within each molecular family and summarized across the 23 families by bootstrap resampling.

### 7.2 Oligothiophene Raman analysis

The $\beta$-n-hexyl 4T, 6T, 8T and 10T oligothiophenes, containing 2, 2, 4 and 4 side chains, respectively, were reconstructed in RDKit from the published molecular structures[41]. For each neutral isolated molecule, 16 ETKDGv3 conformers were generated, and the lowest-energy converged MMFF94s conformer was retained, with UFF used as a fallback. The inter-ring geometry was constrained to a $30°$ twist from the all-anti arrangement. No DFT calculation, basis-set treatment, multiplicity specification or solvent model was used.

SENK-EP used the released QMe14S qme14s composite configuration and benchmark weights, combining ENK-on $H_{ii}$ with EP $H_{ij}$, dedipole and depolar predictions, without oligomer-specific training, fine-tuning or parameter updates.

Projection analysis used normalized mass-weighted modes and two ordered intraring $C_\alpha - C_\beta$ coordinates per thiophene ring. R-like and Z-like patterns used all-positive and alternating signs, respectively, with the Z-like pattern orthogonalized to the R-like pattern. Candidates were restricted to $1350 - 1500\ \text{cm}^{-1}$ and $1485 - 1580\ \text{cm}^{-1}$, respectively. The maximum squared projection-score pair was selected, after which up to five modes accounting for $80\%$ of the squared projection score were retained.

The integrated Raman response was the summed SENK Raman activity of the retained modes. Predicted responses were normalized to 4T for relative comparison with experimental differential Raman cross-sections measured at $1064\ \text{nm}$ for approximately $1\ \text{mM}$ solutions in toluene[41]. Spectra used a Lorentzian width parameter of $12\ \text{cm}^{-1}$, per-trace normalization and vertical offsets. Experimental spectra were digitized from the published spectra[41]. Pearson $r$ was used descriptively for the four 4T-normalized values (n=4). For Figure 7f, the predicted frequency axis was uniformly scaled by a factor of 1.032 for display only; this scaling did not affect the relative R-like Raman-response analysis in Figure 7b-d.

## 8. Spectral Conversion Methods

The predicted response tensors were converted into vibrational spectra by integrating the harmonic-analysis and line-broadening procedure of DetaNet[23]. Specifically, the full Cartesian Hessian was assembled from $H_{ii}$ and $H_{ij}$, symmetrized, mass weighted, and diagonalized to obtain normal-mode frequencies and eigenvectors; IR intensities and Raman activities were then evaluated with the same chain-rule contractions, Raman invariants, Lorentzian broadening kernel, and unit-conversion coefficients as in DetaNet. Relative to the original public implementation, we introduced only minimal numerical-stability safeguards for ill-conditioned predicted Hessians, including non-finite-value cleanup, a tiny diagonal regularization before eigendecomposition, and sign-safe handling of negative eigenvalues prior to filtering non-physical modes. Beyond these stability safeguards, no new spectral post-processing design was introduced.

## Data Availability

All datasets used for model training in this study were obtained from publicly available repositories. The QM9S dataset is available through Figshare at https://figshare.com/articles/dataset/QM9S_dataset/24235333. The QMe14S dataset is available through Figshare at https://figshare.com/s/889262a4e999b5c9a5b3. The SIMG dataset used for stereoelectronic pretraining is available through the Gomes Group repository on Hugging Face at https://huggingface.co/gomesgroup/simg. The qcMol dataset and the corresponding subsets used for electron-prior adaptation are available through the qcMol Download Center at https://structpred.life.tsinghua.edu.cn/qcmol/download.html. The ChEMBL-derived molecular structures used for the external chemical-space analysis were obtained from the Raman-ChEMBL spectral dataset, which is publicly available in two Figshare deposits: https://figshare.com/articles/dataset/Raman-ChEMBL-part1/28593698/2 and https://figshare.com/articles/dataset/Raman-ChEMBL-part2/28594295/2. The R-3B69 structures and reference spectra are publicly available through the MACE-MDP dataset release on Zenodo at https://doi.org/10.5281/zenodo.19205036. No new training dataset was generated in this study.

## Code Availability

The complete source code used in this study is publicly available via GitHub at https://github.com/ztli-ai4s/SENK_main.

## Acknowledgements

This work was supported by the National Natural Science Foundation of China (82425104), the National Key Research and Development Program of China (2022YFC3400501), the Science and Technology Commission of Shanghai Municipality (24JS2830200) and the Shanghai Municipal Education Commission (2024AI01014). The authors acknowledge the ECNU Multifunctional Platform for Innovation (001) for providing computational support.

## Author Contributions

Z. Li designed the SENK model architecture and methodology, performed the model evaluations, interpreted the results, and wrote the manuscript. Z.X. performed the experimental spectroscopic measurements and contributed to the interpretation and discussion of the results. H.F. and J.Y. contributed to the model architecture and methodology. Q.H., Z. Lu and L.X. contributed to the analysis of the model architecture and to the interpretation and theoretical discussion of the results. J.W. reviewed and improved the model architecture, contributed to the interpretation of the results, and revised the manuscript. H.L. conceived and supervised the overall project and revised the manuscript. All authors discussed the results, critically revised and approved the manuscript.

## Competing interests

The authors declare no competing interests.

## References

1. Barone, V. *et al.* Computational molecular spectroscopy. *Nat. Rev. Methods Primer* **1**, 38 (2021).
2. Neese, F. Prediction of molecular properties and molecular spectroscopy with density functional theory: From fundamental theory to exchange-coupling. *Coord. Chem. Rev.* **253**, 526–563 (2009).
3. Kraka, E., Zou, W. & Tao, Y. Decoding chemical information from vibrational spectroscopy data: Local vibrational mode theory. *WIREs Comput. Mol. Sci.* **10**, e1480 (2020).
4. Gastegger, M., Behler, J. & Marquetand, P. Machine learning molecular dynamics for the simulation of infrared spectra. *Chem. Sci.* **8**, 6924–6935 (2017).
5. Ye, S. *et al.* A Machine Learning Protocol for Predicting Protein Infrared Spectra. *J. Am. Chem. Soc.* **142**, 19071–19077 (2020).
6. Ren, H. *et al.* A machine learning vibrational spectroscopy protocol for spectrum prediction and spectrum-based structure recognition. *Fundam. Res.* **1**, 488–494 (2021).
7. Han, R., Ketkaew, R. & Luber, S. A Concise Review on Recent Developments of Machine Learning for the Prediction of Vibrational Spectra. *J. Phys. Chem. A* **126**, 801–812 (2022).
8. Yuan, M., Zou, Z., Luo, Y., Jiang, J. & Hu, W. QMe14S: A Comprehensive and Efficient Spectral Data Set for Small Organic Molecules. *J. Phys. Chem. Lett.* **16**, 3972–3979 (2025).
9. Liang, J., Ling, J., Xu, L. & Zhu, X. A Dataset of Raman and Infrared Spectra as an Extension to the ChEMBL. *Sci. Data* **12**, 939 (2025).
10. Wang, T., Tan, Y., Chen, Y. Z. & Tan, C. Infrared Spectral Analysis for Prediction of Functional Groups Based on Feature-Aggregated Deep Learning. *J. Chem. Inf. Model.* **63**, 4615–4622 (2023).
11. Berger, E., Niemelä, J., Lampela, O., Juffer, A. H. & Komsa, H.-P. Raman Spectra of Amino Acids and Peptides from Machine Learning Polarizabilities. *J. Chem. Inf. Model.* **64**, 4601–4612 (2024).
12. Thomas, M., Brehm, M., Fligg, R., Vöhringer, P. & Kirchner, B. Computing vibrational spectra from ab initio molecular dynamics. *Phys. Chem. Chem. Phys.* **15**, 6608–6622 (2013).
13. Zhang, Y. & Jiang, B. Universal machine learning for the response of atomistic systems to external fields. *Nat. Commun.* **14**, 6424 (2023).

14. Xu, N. *et al.* Tensorial Properties via the Neuroevolution Potential Framework: Fast Simulation of Infrared and Raman Spectra. *J. Chem. Theory Comput.* **20**, 3273–3284 (2024).

15. Ye, S. *et al.* Artificial Intelligence-based Amide-II Infrared Spectroscopy Simulation for Monitoring Protein Hydrogen Bonding Dynamics. *J. Am. Chem. Soc.* **146**, 2663–2672 (2024).

16. Mazzeo, P., Cupellini, L. & Mennucci, B. Multiscale Machine Learning Prediction of Infrared Spectra of Solvated Molecules. *J. Chem. Theory Comput.* **22**, 1883–1895 (2026).

17. Liu, C., Zou, R. & Mo, F. Infrared Spectra Prediction for Functional Group Region Utilizing a Machine Learning Approach with Structural Neighboring Mechanism. *Anal. Chem.* **96**, 15550–15562 (2024).

18. Saquer, N., Iqbal, R., Ellis, J. D. & Yoshimatsu, K. Infrared spectra prediction using attention-based graph neural networks. *Digit. Discov.* **3**, 602–609 (2024).

19. Ji, S. *et al.* A foundational deep learning force field for simulating IR and Raman spectra via molecular dynamics. *Npj Comput. Mater.* (2026).

20. Han, B. *et al.* AI-powered exploration of molecular vibrations, phonons, and spectroscopy. *Digit. Discov.* **4**, 584–624 (2025).

21. Westermayr, J. & Marquetand, P. Machine learning spectroscopy to advance computation and analysis. *Chem. Sci.* **16**, 21660–21676 (2025).

22. Schütt, K., Unke, O. & Gastegger, M. Equivariant message passing for the prediction of tensorial properties and molecular spectra. in *Proceedings of the 38th International Conference on Machine Learning* 9377–9388 (PMLR, 2021).

23. Zou, Z. *et al.* A deep learning model for predicting selected organic molecular spectra. *Nat. Comput. Sci.* **3**, 957–964 (2023).

24. Yang, X., Zhang, X., Zhang, Y., Jiang, J. & Hu, W. Deep Learning Protocol for Predicting Full-Spectrum Infrared and Raman Spectra of Polypeptides and Proteins Using All-Atom Models. *J. Phys. Chem. Lett.* **16**, 2023–2028 (2025).

25. Xu, Y. *et al.* Pretrained E(3)-equivariant message-passing neural networks with multi-level representations for organic molecule spectra prediction. *Npj Comput. Mater.* **11**, 203 (2025).

26. Chen, Y., Pios, S. V., Gelin, M. F. & Chen, L. Accelerating Molecular Vibrational Spectra Simulations with a Physically Informed Deep Learning Model. *J. Chem. Theory Comput.* **20**, 4703–4710 (2024).

27. Berger, E. & Komsa, H.-P. Polarizability models for simulations of finite temperature Raman spectra from machine learning molecular dynamics. *Phys. Rev. Mater.* **8**, 043802 (2024).

28. Fang, M. *et al.* Transferability of Machine Learning Models for Predicting Raman Spectra. *J. Phys. Chem. A* **128**, 2286–2294 (2024).

29. Gönnheimer, N., Reuter, K., Kapil, V. & Margraf, J. T. MACE-MDP: A General Dipole and Polarizability Model for Organic Molecules and Materials. *ChemRxiv* **2026**,.

30. Chen, K. & Luber, S. Field-aware and charge-informed machine learning for predicting molecular and condensed-phase responses and vibrational spectra. *Npj Comput. Mater.* (2026).

31. Schienbein, P. Mimyria: Machine-Learned Vibrational Spectroscopy for Aqueous Systems Made Simple. *J. Chem. Theory Comput.* **22**, 4626–4640 (2026).

32. Lazzaroni, P., Sharma, S. & Rossi, M. Investigating anharmonicities in polarization-orientation Raman spectra of acene crystals with machine learning. *Phys. Rev. B* **113**, 054309 (2026).

33. Zou, W., Tao, Y., Freindorf, M., Cremer, D. & Kraka, E. Local vibrational force constants – From the assessment of empirical force constants to the description of bonding in large systems. *Chem. Phys. Lett.* **748**, 137337 (2020).

34. Prodan, E. & Kohn, W. Nearsightedness of electronic matter. *Proc. Natl. Acad. Sci.* **102**, 11635–11638 (2005).

35. Fias, S., Heidar-Zadeh, F., Geerlings, P. & Ayers, P. W. Chemical transferability of functional groups follows from the nearsightedness of electronic matter. *Proc. Natl. Acad. Sci.* **114**, 11633–11638 (2017).

36. Baiz, C. R. *et al.* Vibrational Spectroscopic Map, Vibrational Spectroscopy, and Intermolecular Interaction. *Chem. Rev.* **120**, 7152–7218 (2020).

37. Spencer, R. J., Zhanserkeev, A. A., Yang, E. L. & Steele, R. P. The Near-Sightedness of Many-Body Interactions in Anharmonic Vibrational Couplings. *J. Am. Chem. Soc.* **146**, 15376–15392 (2024).

38. Foster, J. P. & Weinhold, F. Natural hybrid orbitals. *J. Am. Chem. Soc.* **102**, 7211–7218 (1980).

39. Reed, A. E., Weinstock, R. B. & Weinhold, F. Natural population analysis. *J. Chem. Phys.* **83**, 735–746 (1985).

40. Weinhold, F. Natural bond orbital analysis: A critical overview of relationships to alternative bonding perspectives. *J. Comput. Chem.* **33**, 2363–2379 (2012).

41. Huff, G. S., Gallaher, J. K., Hodgkiss, J. M. & Gordon, K. C. No single DFT method can predict Raman cross-sections, frequencies and electronic absorption maxima of oligothiophenes. *Synth. Met.* **231**, 1–6 (2017).

42. Řezáč, J., Huang, Y., Hobza, P. & Beran, G. J. O. Benchmark Calculations of Three-Body Intermolecular Interactions and the Performance of Low-Cost Electronic Structure Methods. *J. Chem. Theory Comput.* **11**, 3065–3079 (2015).

43. Du, J. *et al.* Raman-guided subcellular pharmaco-metabolomics for metastatic melanoma cells. *Nat. Commun.* **11**, 4830 (2020).

44. Lu, T. & Chen, F. Multiwfn: A multifunctional wavefunction analyzer. *J. Comput. Chem.* **33**, 580–592 (2012).

45. Lu, T. A comprehensive electron wavefunction analysis toolbox for chemists, Multiwfn. *J. Chem. Phys.* **161**, 082503 (2024).

46. Boiko, D. A., Reschützegger, T., Sanchez-Lengeling, B., Blau, S. M. & Gomes, G. Advancing molecular machine learning representations with stereoelectronics-infused molecular graphs. *Nat. Mach. Intell.* **7**, 771–781 (2025).

47. Wang, H., Zhang, Z. & Gong, H. A dataset of 1.2 million molecules with DFT-level quantum chemical annotations for molecular representation learning. *Commun. Chem.* **9**, 278 (2026).

48. Frisch, M. J. *et al.* Gaussian 16 Revision B.01. (2016).

49. Stephens, P. J., Devlin, F. J., Chabalowski, C. F. & Frisch, M. J. Ab Initio Calculation of Vibrational Absorption and Circular Dichroism Spectra Using Density Functional Force Fields. *J. Phys. Chem.* **98**, 11623–11627 (1994).

50. Weigend, F. & Ahlrichs, R. Balanced basis sets of split valence, triple zeta valence and quadruple zeta valence quality for H to Rn: Design and assessment of accuracy. *Phys. Chem. Chem. Phys.* **7**, 3297–3305 (2005).

51. Goodfellow, A. S. & Nguyen, B. N. Graph-Based Internal Coordinate Analysis for Transition State Characterization. *J. Chem. Theory Comput.* **22**, 2348–2357 (2026).

52. Liao, Y.-L., Wood, B., Das, A. & Smidt, T. EquiformerV2: Improved Equivariant Transformer for Scaling to Higher-Degree Representations. in *International Conference on*

*Learning Representations* (2024).

53. Liao, Y.-L. & Smidt, T. Equiformer: Equivariant Graph Attention Transformer for 3D Atomistic Graphs. in *International Conference on Learning Representations* (2023).

54. Wojke, N., Bewley, A. & Paulus, D. Simple online and realtime tracking with a deep association metric. in *2017 IEEE International Conference on Image Processing (ICIP)* 3645–3649 (2017).

55. Batselier, K., Chen, Z. & Wong, N. A Tensor Network Kalman filter with an application in recursive MIMO Volterra system identification. *Automatica* **84**, 17–25 (2017).

56. Holtmann, T., Stenger, D., Posada Moreno, A. F., Solowjow, F. & Trimpe, J. S. Sailing Towards Zero-Shot State Estimation using Foundation Models Combined with a UKF. in *2025 IEEE 64th Conference on Decision and Control (CDC)* (IEEE, 2025).

57. An, J. *et al.* Equivariant Masked Position Prediction for Efficient Molecular Representation. in *International Conference on Learning Representations* (2025).